\documentclass[sigconf]{acmart}

\usepackage[utf8]{inputenc} 
\usepackage[T1]{fontenc}    
\usepackage{hyperref}       
\usepackage{url}            
\usepackage{booktabs}       
\usepackage{amsfonts}       
\usepackage{nicefrac}       
\usepackage{microtype}      
\usepackage{graphicx}
\usepackage{xcolor}
\usepackage[ruled,vlined]{algorithm2e}
\usepackage{subcaption}
\usepackage{caption}
\usepackage{multirow}
\usepackage{float}
\usepackage{varwidth}
\usepackage{balance}
\AtBeginDocument{%
  \providecommand\BibTeX{{%
    \normalfont B\kern-0.5em{\scshape i\kern-0.25em b}\kern-0.8em\TeX}}}

\acmDOI{}

\acmConference[WSDM ConvERSe'20]{WSDM 2020 Workshop on Conversational Systems for E-Commerce Recommendations and Search }{Feb 07, 2020}{Houston, TX}

\begin{document}

\title{MT-BioNER: Multi-task Learning for Biomedical Named Entity Recognition using Deep Bidirectional Transformers}

\author{Muhammad Raza Khan}
\authornote{The first two authors contributed equally to this paper.}
\affiliation{%
	\institution{Azure Text Analytics, Microsoft Cloud \& AI}
}
\email{razakhan@microsoft.com}
\author{Morteza Ziyadi}
\authornotemark[1]
\affiliation{%
	\institution{Microsoft Dynamics 365 AI}
}
\email{morteza.ziyadi@microsoft.com}

\author{Mohamed AbdelHady}
\authornote{This work was done while the author was a data science manager in the Azure AI group at Microsoft.}
\affiliation{%
	\institution{Alexa AI, Amazon}
}
\email{mbdeamz@amazon.com}

%
%


\begin{abstract}
Conversational agents such as Cortana, Alexa and Siri are continuously working on increasing their capabilities by adding new domains. The support of a new domain includes the design and development of a number of NLU components for domain classification, intents classification and slots tagging (including named entity recognition). Each component only performs well when trained on a large amount of labeled data. Second, these components are deployed on limited-memory devices which requires some model compression. Third, for some domains such as the health domain, it is hard to find a single training data set that covers all the required slot types. To overcome these mentioned problems, we present a multi-task transformer-based neural architecture for slot tagging. We consider the training of a slot tagger using multiple data sets covering different slot types as a multi-task learning problem. The experimental results on the biomedical domain have shown that the proposed approach outperforms the previous state-of-the-art systems for slot tagging on the different benchmark biomedical datasets in terms of (time and memory) efficiency and effectiveness. The output slot tagger can be used by the conversational agent to better identify entities in the input utterances.
\end{abstract}





\maketitle

\section{Introduction}
Healthcare industry is going through a technological transformation especially through the increasing adoption of conversational agents including voice assistants (such as Cortana, Alexa, Google Assistant,  and Siri) and the medical chatbots \footnote{https://www.usertesting.com/about-us/press/press-releases/healthcare-chatbot-apps-on-the-rise-but-cx-falls-short}. These conversational agents hold great potential when it comes to transforming the way how consumers connect with the providers but recent reports suggest that these conversational agents need to improve their performance and usability in order to fulfill their potential. 

The field of biomedical text understanding has received increased attraction in the last few years. With the increase accessibility of the medical records, scientific reports and publications along with the better tools and algorithms to process these big datasets, precision medicine and diagnosis have the potential to make the treatments much more effective. In addition, the relations between different symptoms and diseases, side-effects of different medicines can be more accurately identified based on text mining on these big datasets. The performance of the biomedical text understanding systems depends heavily on the accuracy of the underlying Biomedical Named Entity Recognition component (BioNER) i.e., the ability of these systems to recognize and classify different types of biomedical entities/slots present in the input utterances. 

Slot tagging and Named Entity Recognition (NER) extract semantic constituents by using the words of an input sentence/utterance to fill in predefined slots in a semantic frame. The slot tagging problem can be regarded as a supervised sequence labeling task, which assigns an appropriate semantic label to each word in the given input sentence/utterance. The slot type could be a quantity (such as time, temperature, flight number, or building number) or a named entity such as person or location. One of the key challenges these text understanding systems face is better identification of symptoms and other entities in their input utterances. In other words, these conversational agents need to improve their slot tagging capabilities\cite{chatbot_report}. In general, conversational agents ability to perform different tasks like topic classification or intent detection depends heavily on their ability to accurately identify slot types in the input and the researchers have been trying to both improve the named entity recognition component then use the improved component for the downstream task. Some example of this trend include Entity-aware topic classification \cite{ahmadvand2019concet}, de-identification of medical records for conversational systems \cite{cohn2017deid} and improving inquisitiveness of conversational systems through NER \cite{reshmi2018enhancing}. 

Building an NER component with high precision and recall is technically challenging because of some reasons: 1) Requirement of hand-crafted features for each of the label to increase the system performance, 2) Lack of extensively labelled training dataset. 3) For conversational agents, the slot tagger may be deployed on limited-memory devices which requires model compression or knowledge distillation to fit into memory. 4) For some domains such as the health/biomedical domain, it is hard to find a single training data set that covers all the required slot types. 

Many of the current state of the art BioNER systems rely on hand-crafted features for each of the labels in the data. The computation of these hand-crafted features takes most of the computation time. Furthermore, the usage of these hand-crafted features results in a system that is optimized for the training dataset. Recent advances in the neural network-based feature learning approaches have helped researchers to develop BioNER systems that are no longer dependent on the manual feature engineering process. However, the performance of these neural network-based techniques depends heavily on the presence of big and high-quality labelled training dataset. Such large datasets can be very difficult to obtain in the biomedical domain because of the cost and privacy issues.

Multi-task learning is one way to solve the problem of lack of extensive training data. Basic premise behind multi-task learning is that different datasets may have semantic and syntactic similarities and these similarities can help in training of a much more optimized model as compared to the one trained on a single dataset. It also helps to reduce model overfitting. However, multi-task learning models without pre-trained weights can take a long time to train on large datasets. In contrast, pre-trained language model based approaches (\cite{devlin2018bert}; \cite{peters2018deep}; and \cite{radford2018improving}) combined with multi-task learning have recently started showing promising results (\cite{liu2019multi}. 

To overcome these challenges, we present a multi-task transformer-based neural architecture for slot tagging and applied it to the Biomedical domain (MT-BioNER). We consider the training of a slot tagger using multiple data sets covering different slot types as a multi-task learning problem. The experimental results on the biomedical domain have shown that the proposed approach outperforms the previous state-of-the-art methods for slot tagging on the different benchmark biomedical datasets in terms of (time and memory) and effectiveness. These results can be used by the biomedical conversational agent to better identify entities in the input utterances.

\begin{figure*}[!htb]
	\centering
	\includegraphics[width=\linewidth]{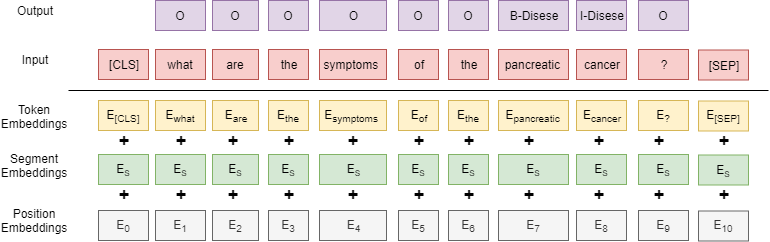}
	\caption{\textbf{The representation of MT-BioNER input sequence and output label sequence. The input embeddings are the sum of the token embeddings, the segmentation embeddings and the position embeddings}}
	\label{fig:data-format}
\end{figure*}

\section{Related Work}
\label{RelatedWork}

Many recent studies on BioNER have used neural networks to overcome the requirement of hand-crafted feature generation. \cite{crichton2017neural} used a word and its surrounding context as input to the convolutional neural network while \cite{habibi2017deep} used word embeddings as input to the BiLSTM-CRF model for named entity recognition. Both of these models are based on single task / dataset approaches and they cannot use the information contained in multiple datasets as it is done in multi-task learning (MTL).
Although the multi-task learning (MTL) based approaches have been widely used in the NLP research (for example, \cite{collobert2008unified} used it for standard NLP tasks like POS Tagging, Chunking etc.,), but application of MTL in biomedical text mining has not seen promising results primarily because many of the approaches (e.g. \cite{crichton2017neural})  used word level features as input ignoring sub-word information which can be quite important in the biomedical domain. To overcome this challenge, \cite{wang2018cross} used a multi-task BiLSTM-CRF model with an additional context dependent BiLSTM layer for modelling character sequences and was able to beat the benchmark results on five canonical biomedical datasets. Our proposed system is different from their scheme as we are able to combine the advantages of both the multi-task learning and pre-trained language models. Furthermore, the use of character-based LSTM models can be slower in terms of training and scoring time. Actual comparison of our model and \cite{wang2018cross} work is described later on in this paper.
Pre-trained language model based approaches have been popular for different biomedical text mining tasks. For instance, \cite{sachan2017effective} used transfer learning based approach to pretrain weights of an LSTM by training it in both  forward and reverse direction using Word2Vec (\cite{mikolov2013distributed}) embedding trained on a large collection of biomedical documents. In such approaches, Word2vec model needs be fine-tuned according to the variations in the biomedical data (\cite{Pyysalo:2013b}).
Recent developments of ELMO (\cite{peters2018deep}); GPT (\cite{radford2018improving}); and BERT (\cite{devlin2018bert}) language models have proven the effectiveness of the contextualized representations for different NLP tasks. These contextual representation learning models fine tune unsupervised objectives on text data. For instance, BERT (\cite{devlin2018bert}) uses multi-layer bidirectional Transformer on plain text for masked word prediction and next sentence prediction tasks. However, these unsupervised models must be fine-tuned to achieve better results for the specific prediction tasks using additional task-specific layers and datasets. The examples of BERT based models fine-tuned for biomedical domain include:
\begin{enumerate}
\item BioBERT \cite{lee2019biobert} that fine-tunes BERT model on Pubmed abstracts and PMC full text articles along with English Wikipedia and Books corpus
\item ClinicalBERT \cite{huang2019clinicalbert} which fine-tunes the BERT model for the purpose of hospital re-admission prediction
\item \citet{alsentzer2019publicly} fine-tuned BERT/BioBERT models for the purpose of named entity recognition on a single dataset (task) and deidentification on clinical texts.
\end{enumerate}

We can see two clear trends in the related research literature described in this section: 1) Multi-task based learning approaches that are able to use the sub-words based information are able to beat the other models for biomedical named entity recognition. 2) Contextual representation models like BERT have been quite successful for language modeling related problems. Based on these observations, we see a lot of potential for the approaches that combine both the multi-task learning and language model pretraining based approaches and this paper presents one such approach.  Our approach takes inspiration from the recent multi-task deep neural network (MT-DNN, \cite{liu2019multi} work which combines both multi-task learning and BERT language model, but MT-DNN is optimized for general natural language understanding (NLU) tasks. In contrast, our model MT-BioNER is optimized for biomedical named entity recognition using BioBERT as the shared layers and the different data sets in the task-specific layers.

\section{MT-BioNER}
\begin{figure}[!t]
	\centering
	\includegraphics[width=\linewidth]{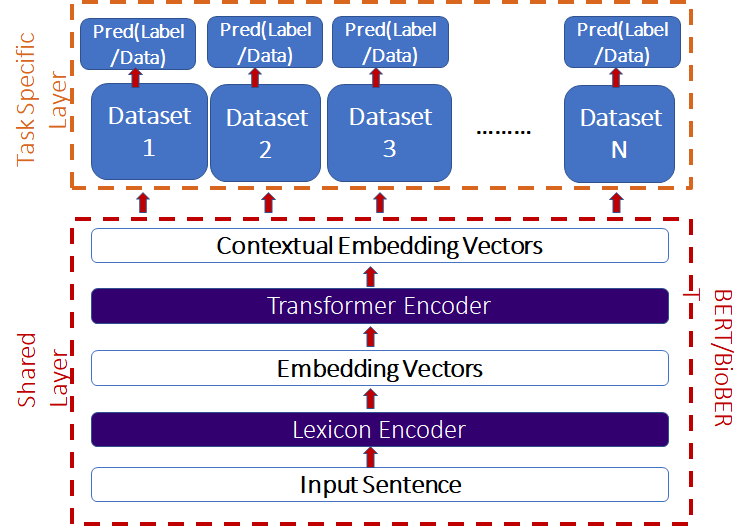}
	\caption{\textbf{Architecture of the MT-BioNER model.}}
	\label{fig:arch}
\end{figure}

\subsection{Model Architecture}
As described in the introduction and related work sections, our model combines pre-trained language models (using BERT as the shared layers) and transfer learning (using task specific output layers). Figure \ref{fig:arch} shows the architecture of the MT-BioNER model. The lower layers are shared across all tasks/datasets while the top layers are specific for each dataset (entity types). The input sentence is first represented as a sequence of embedding vectors ($X$), one for each token which consists of word, position and segment embeddings. Then the Transformer encoder captures the contextual information for each token and generates the shared contextual embedding vectors. Finally, for each task/dataset, a shallow layer is used to generate task-specific representations, followed by operations necessary for entity recognition. The input sentence is first represented as a sequence of embedding vectors ($X$), one for each token. Then the Transformer encoder captures the contextual information for each token and generates the shared contextual embedding vectors. Finally, for each task/dataset, a shallow layer is used to generate task-specific representations, followed by operations necessary for entity recognition.  
The input word sequence and output label sequence are represented as shown in Figure \ref{fig:data-format}. The model architecture details are as follows:
\begin{itemize}
\item {\bf Lexicon Encoder layer}: The input sentence $s = \{w_1, ..., w_n\}$ is a sequence of tokens of length $n$. Note that two additional labels $[CLS]$ and $[SEP]$ are used to represent the start and end of a sentence, respectively. The first token $w_1$ is always the $[CLS]$ token and the last token is $[SEP]$. The lexicon encoder maps $s$ into a sequence of input embedding vectors $X = \{x_1, ..., x_n\}$ constructed by concatenating the corresponding word, segment, and position embeddings  to produce the input to the Transformer Encoder layers. 

\item {\bf Transformer Encoder layers}: The encoder of the multilayer bidirectional Transformer (\cite{vaswani2017attention}) is used to map the input representation vectors into contextual embedding vectors. In our experiments, we use BERT encoder model \cite{devlin2018bert} as the shared layers across different tasks (datasets). We fine-tune the BERT model alongside the task-specific layers during the training phase using a multi-task objective function.

\item {\bf Task-Specific Layers}: We use a shallow linear layer for each of the tasks/datasets. Depending on the slot types covered by each training dataset, we treat each dataset as a separate slot tagging task and add a separate output linear layer as the last layer of the network to learn the entities in that dataset. In our experiments, we use softmax layer. We have conducted additional experiments where we replaced the softmax layer with two layers: a feedforward layer and a CRF layer. But the training takes longer and we didn't get significant improvement in the test accuracy.
\end{itemize}

\begin{algorithm}[!t]
	
	\caption{Training Algorithm I\label{algo:1}}
	\BlankLine
	Initialize model parameters $\theta$\\
	$\vert$ shared layers parameter $\theta_i^{BERT}$ by BERT\\
	$\vert$ task-specific layer parameters $\theta_i^o$ randomly\\
	Set max epochs: $epoch_{max}$
	
	\BlankLine
	\ForEach {$epoch$ in {1,2, $ epoch_{max}$}}{
		
		Merge all the datasets: $D= D_1 U D_2 .... U D_T$\\
		Shuffle D
		\ForEach {$b_t$ in $D$}{
			//b\_t is a mini-batch of task t.  \\
			Computer Loss L($\theta$) (Eq 1)	\\
			Compute gradient: $\nabla (\theta)$\\
			Update model: $\theta = \theta - \eta \nabla \theta$		
		}	
	}
\end{algorithm}

\subsection{Training}
We use a similar objective function as in \cite{wang2018cross} to train the multi-task model. The formal definition of the multi-task setting is as follows. Given $m$ datasets, for $i \in \{1, ..., m\}$, each dataset $D_i$ consists of $n_i$ training samples, i.e., $D_i = \{(s^i_j, y^i_j)\}^{ni}_{j=1}$. We denote the training set for each dataset $D_i$ as $X_i = \{X^i_1 , ..., X^i_{ni} \}$ where $X^i_j= \{X^i_{j1},X^i_{j2}, \dots, X^i_{jn}\}$ is the sequence of feature representations of the input word sequence $s^i_j$ of length $n$. The set of labels for each dataset is $Y_i = \{y^i_1, ..., y^i_{ni}\}$ where $y^i_j$ is the output label sequence of $s^i_j$. The multi-task model consists of $m$ different models each trained on separate dataset while sharing part of the model parameters across datasets. The loss function $L$ of the multi-task model is 

\begin{equation}
L =\sum_{i=1}^{m} \lambda_i L_i(\theta_{i}^{BERT}, \theta_{i}^{o}) = \sum_{i=1}^{m}  \lambda_i\log P(Y_i|X_i; \theta_{i}^{BERT},\theta_{i}^{o})
\label{eq:1}
\end{equation}

The training maximizes the log-likelihood $P(Y_i|X_i; \theta_{i}^{BERT},\theta_{i}^{o})$ of the label sequence $y$ given the input sequence $X$ for each given training data set $D_i$ as shown in Eq. \ref{eq:1} where the cross-entropy loss is used as the loss function. Cross-entropy loss increases as the predicted probability diverges from the actual label. The contribution of each dataset $D_i$ is controlled by the hyperparameter $\lambda_i$. In our experiments, we assume that all the data sets have the same contribution and set $\lambda_i=1$ for all datasets. The BERT shared layer and the $i^{th}$ task-dependent layer parameters are represented by $\theta_{i}^{BERT}$ and $\theta_{i}^{o}$, respectively. We conducted two variants of transfer learning. First, we freeze the shared layer and only fine-tune the task-specific layers (that is, $\theta_{i}^{BERT} = \theta_{initial}^{BERT}$ for all datasets). In the second variant, we fine-tune the whole network. (That is, $\theta_{i}^{BERT} = \theta_{tuned}^{BERT}$).

We use stochastic gradient descent (SGD) to learn the parameters of all shared layers and task-specific layers as shown in Algorithm \ref{algo:1} (based on \cite{liu2019multi}).  We initialize the shared layers with the pre-trained BERT model while the task-dependent layers are initialized randomly. After creating mini-batches of each dataset, we combine all the datasets and shuffle the mini-batches. At each epoch, we train the model on all the mini-batches and then at each batch-iteration, a mini-batch $b_t$ corresponding to task $t$ is selected (from all the datasets), and the model is updated according to the objective for the task $t$.

\begin{tabular}[!htb]{c}
	
	\begin{minipage}{.48\textwidth}
		\begin{tabular}[!t]{llp{2.5cm}}
			
			\midrule
			\textbf{Dataset}     & \textbf{Size}     & \textbf{Entity Type and Counts} \\
			\midrule
			BC2GM	&20,000 sentences	&Gene/Protein (24,583)\\
			BC5CDR	&1500 articles	&Chemical (15,935), Disease (12,852)\\
			NCBI-Disease&	793 abstracts	&Disease (6,881)\\
			JNLPBA	& 2,404 abstracts	&Gene/ Protein (35,336), Cell Line (4,330), DNA (10,589), Cell Type (8,649), RNA (1,069)\\
			\bottomrule
			
		\end{tabular}
		\captionof{table}{Datasets used in the experiments
			\normalfont{All datasets can be downloaded form \url{https://github.com/cambridgeltl/MTL-Bioinformatics-2016}}
		}	\label{Table:1}
	\end{minipage}
\end{tabular}

\section{Experiments}
\subsection{Data Preparation}
We evaluate the performance of the proposed approach on four benchmark datasets used by \citet{sachan2017effective}. Table \ref{Table:1} gives a summary of these datasets based on the number of sentences, and entities. We used these publicly available datasets in order to make the experiments reproducible. We use the same training, development and test sets splits according to \citet{crichton2017neural} for each dataset. As in \citet{wang2018cross} and \citet{sachan2017effective}, we use training and development sets to train the final model.
As part of the data preprocessing step, word labels are encoded using an IOB scheme. In this scheme, for example, a word describing a disease entity is tagged with "B-Disease" if it is at the beginning of the entity, and "I-Disease" if it is inside the entity. All other words not describing entities of interest are tagged as 'O'.

\begin{table*}[!htb]

	\centering
	\begin{tabular}{p{1.45cm}l|p{1.2cm}p{1.2cm}p{1.1cm}|lp{1.42cm}p{1.4cm}}
		\toprule
		Dataset&Metric & \multicolumn{3}{c}{Single-Task Learning}& \multicolumn{3}{c}{Multi-Task Learning} \\
		&      & Benchmark &BiLSTM &BiLM & MTM-CW & \multicolumn{2}{c}{MT-BioNER}\\
		\midrule
		& & & & & & 3-datasets&4-datasets\\
		\midrule
		\multirow{3}{1.2cm}{BC2GM (Genes)}	&Precision	&-	&81.57	&81.81	&\textbf{82.10}	&82.01	&80.33\\
		&Recall	&-	&79.48	&81.57	&79.42	&\textbf{84.04}	&82.82\\
		&F1	&-	&80.51	&81.69	&80.74	&\textbf{83.01}	&81.55\\
		\midrule
		\multirow{3}{1.2cm}{BC5CDR (Chemical)}
		& Precision	& \textbf{89.21}& 	87.60	& 88.10	& 89.10	& 88.46	& 87.99\\
		& Recall	& 84.45	& 86.25	& 90.49	& 88.47	& \textbf{90.52}	& 90.16\\
		& F1& 	86.76	& 86.92	& 89.28	& 88.78	& \textbf{89.5}	& 89.06\\
		\midrule
		\multirow{3}{1.2cm}{NCBI (Disease)}	&Precision	&85.10	&86.11	&86.41	&85.86	&\textbf{86.73}	&84.50\\
		&Recall	&80.80	&85.49	&88.31	&86.42	&\textbf{89.7}	&88.98\\
		&F1	&82.90	&85.80	&87.34	&86.14	&\textbf{88.1}	&86.68\\
		\midrule
		\multirow{3}{1.2cm}{JNLPBA	(Genes etc.)}&Precision	&69.42	&71.35	&\textbf{71.39}	&70.91	&-	&67.40\\
		&Recall	&75.99	&75.74	&79.06	&76.34	&-	&\textbf{79.35}\\
		&F1	&72.55	&73.48	&\textbf{75.03}	&73.52	&-	&72.89\\
		\bottomrule
	\end{tabular}	
	\caption{Comparison of MT-BioNER model and recent state-of-the-art models.\normalfont{Source of benchmark performance scores of datasets are: NCBI-disease: \cite{leaman2016taggerone}; BC5CDR: \cite{li2015hitsz_cdr}; JNLPBA: \cite{guodong2004exploring}; MTM-CW was proposed by \citep{wang2018cross}; BiLSTM was proposed by \citep{habibi2017deep}. The performance scores for these NER models are referred from \citep{sachan2017effective}}}
	\label{Table:results}
\end{table*}

\subsection{Training and Evaluation Details}
We test our method on four benchmark datasets used by \cite{sachan2017effective} . All the neural network models are trained on one Tesla K80 GPU using PyTorch framework . To train our neural models, we use BertAdam optimizer with a learning rate of 5e-5  and a linear scheduler with a warmup proportion of 0.4, and a weight decay of 0.01 applied to every epoch of training. We use a batch size of 32, and the maximum sentence length of 128 words. We use BioBERT model (\cite{lee2019biobert}) as the domain-specific language model. Lee et al. \cite{lee2019biobert} also presented the use of BioBERT for biomedical NER scenario. But their scheme is to develop different models for different datasets.

\section{Results}
\subsection{Performance}
We compare our proposed MT-BioNER model with state-of-the-art BioNER systems such as  the single task LSTM-CRF model of \citet{habibi2017deep} (BiLSTM), the multi-task model of \citet{wang2018cross} (MTM-CW), and transfer learning approach of \citet{sachan2017effective} (BiLM-NER). We show the precision, recall, and F1 scores of the models trained on three and four datasets in Table \ref{Table:results}. BioBERT model is used as the shared layers for these results. From the results, we see that our approach trained on three datasets obtains the maximum recall and F1 scores. We should mention that our model is based on multi-task approach and achieves better performance even compared to \citet{sachan2017effective} single task transfer learning approach (BiLM-NER) in which the whole network is only trained on a single dataset. Moreover, our model trained on four datasets performs better on recall and F1-score compared to MTM-CW approach which is a multi-task model. Another interesting result is that our model achieves the highest recall among all the other approaches. But it has a lack in precision score. To further improve the precision, we can add dictionaries as features which could be an interesting future work. Also, it shows the potential capability of BERT language model to provide the semantic feature representations for the multi-task NER tasks. 

\subsection{Training/Scoring Time and Model Size}
As mentioned before, all the neural network models are trained on one Tesla K80 GPU.  We compare the training time of our model with \citet{wang2018cross} as we utilize it for benchmark comparison, and they have made their model publicly available. We find that for four datasets, it takes on average ~40 minutes per epoch (1.45 sec/mini-batch, with total of 1,537 minibatches, and batch size of 32)  and with a total number of 8 epoch, the full training on 4 datasets is less than 6 hours. On average it is equivalent to 0.36 sec per sentence to train a final model which compare to \citet{wang2018cross} it is at least twice faster to train our model. One of the main reasons is that \citet{wang2018cross} model uses character, and word level bidirectional LSTM layers which could take more time to train a model. 

\begin{tabular}[!htb]{c}
	\begin{minipage}{.48\textwidth}
		\begin{tabular}[t]{p{1.cm}p{1.1cm}p{1.7cm}p{1.2cm}}	
			\midrule
			Dataset     & Metric     & BERT-Base & BioBERT \\
			\midrule
			&	Epochs	&50	&8\\
			\midrule
			BC2GM	&Precision	&77.86	&80.33\\
			&Recall	&80.57	&82.82\\
			&F1	&79.19	&81.55\\
			\midrule
			BC5CDR&		Precision	&85.92	&87.99\\
			&Recall&	87.79&	90.16\\
			&F1&	86.85&	89.06\\
			\midrule
			NCBI	&Precision	&84.78	&84.50\\
			&Recall	&86.90	&88.98\\
			&F1	&85.83	&86.68\\
			\midrule
			JNLPBA	&Precision	&65.40	&67.40\\
			&Recall	&75.58	&79.35\\
			&F1	&70.12	&72.89\\
			\bottomrule
		\end{tabular}
		\captionof{table}{Domain Adaptation Experiment}	\label{table:3}
	\end{minipage}
\end{tabular}

As another comparison, we study the scoring (inference) time as well since in the real-world applications, this parameter plays an important role. For the scoring time, we test it on a single dataset of BC5CDR test set and it takes ~80 seconds to run the prediction which compare to \citet{wang2018cross} model, it is at least twice faster and we think the character level LSTM layers could be the main reason. We should also mention that prediction time of a multi-task model in general is faster than multiple single-task models such as \citet{sachan2017effective} since we can run the shared layers once and the remaining shallow-task dependent layers could be run in parallel.
Model size is another factor in real-world applications. Using our approach, the model size is ~430 MB which is bigger than \cite{wang2018cross} model which is ~220MB. Since we are using BERT base model with 12 neural network layers, the bigger size of the model was expected. Moreover, compared to single-task models which require multiple models to identify multiple entities (form different datasets), multi-task approaches have the advantage of providing a single model for all the entity types across different datasets which is a big advantage in real-world production environments. 

\subsection{Fine-tuning Approaches Study}
To achieve the best results, we fine-tune the shared language model alongside the task-specific layers during training phase. An interesting study could be to investigate if we only train the task-dependent layers and freeze the language model to its pre-trained weights. We run this experiment with the same parameters as other experiments and observe a poor performance of the model, i.e., for all the datasets the F1-score is less than 60\%. This poor performance could be explained as following. Since we are training only the task-dependent layers, the shared layers act as fixed embedding layer which is not trainable. So, the whole  network to train would be a shallow linear layer and using a linear layer with an embedding in a multi-task fashion may not give good results as it is observed in this study.

\subsection{Domain Adaptation Study}
The benchmark results shown in the Table 2 are achieved based on BioBERT as the shared layer which is in-domain language model. To analyze the domain adaptation scenario, we run the same experiment on the general BERT base model. In the experiments, we use BERT-base-Cased model instead of BioBERT since BioBERT is a fine-tuned version of BERT-base-Cased model. Table 3 shows the performance comparison of these two scenarios. We observe that BERT base model reaches to a margin of state-of-the-art results, but it requires a greater number of training iterations to adapt to a new domain. To mention that in order to further improve these results, one might first fine-tune the BERT language model using all datasets and save a new language model and then use this new model in our approach. This approach is very similar to the work of Sachan et al. (\cite{sachan2017effective}) on transfer learning in BiLM-NER in which they have created their own language model based on character-level CNN layer with word embedding layers as lexicon encoder and bidirectional LSTM layers to extract the contextual embedding vectors. We didn’t experiment this approach in this paper since we focused on the multi-task study of the work, but it could be an interesting experiment for future works.   

%

\begin{figure}[!htb]
	\centering
	\begin{subfigure}{.5\textwidth}
		\centering
		\includegraphics[width=\linewidth, height=1.9in]{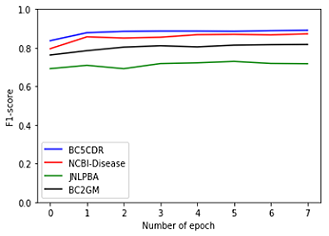}
		\caption{Performance of Training Algorithm I}
	\end{subfigure}
	\hfill
	\begin{subfigure}{.5\textwidth}
		\centering
		\includegraphics[width=\linewidth, height=1.9in]{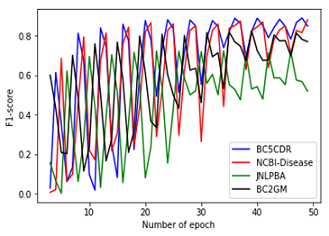}
		\caption{Performance of Training Algorithm II}
	\end{subfigure}
	\caption{Comparison of Algorithm I and II}
	\label{fig:2}
\end{figure}

\subsection{Training Schema Study}
To achieve the best results, we utilized the training scheme of MT-DNN (\citet{liu2019multi}) work, in which, at each training iteration (epoch), we train on all the batches of all the datasets by selecting a random batch from all datasets.  In \citet{wang2018cross}, they utilize a different training scheme in which at each iteration a random dataset is selected, and different batches of that specific dataset is fed into the pipeline to train. Using this scheme, they run several hundreds of epochs to train the model and it makes sense since they train the language model part of their approach as well. But, in our experiment, we need a smaller number of iterations due to availability of the pre-trained language model. Thus, we run similar experiment but instead of random selection, we iterate on the order of the datasets (Algorithm \ref{algo:2}). 
Figure \ref{fig:2} show the performance of training algorithms over the number of iterations for algorithm I, and II, respectively. It is clear that compared to algorithm I which achieve the best results with less number of iterations, training scheme II is performing poor and the learning process is not stable since the model gets biased on the latest dataset that it is trained on.

\begin{algorithm}[!ht]
	
	\caption{Training Algorithm II\label{algo:2}}
	\BlankLine
	Initialize model parameters $\theta$\\
	$\vert$ shared layers parameter $\theta_i^{BERT}$ by BERT\\
	$\vert$ task-specific layer parameters $\theta_i^o$ randomly\\
	Set max epochs: $epoch_{max}$
	
	\ForEach{$t$ in {1,2,..., $T$}}{Pack the dataset t into mini-batch: $D_t.$ }
	\ForEach {$epoch$ in {1,2, $ epoch_{max}$}}{
		
		Select dataset $D_t$ based on the taskorder \\
		\ForEach {$b_t$ in $D$}{
			//b\_t is a mini-batch of task t.  \\
			Computer Loss L($\theta$) (Eq 1)	\\
			Compute gradient: $\nabla (\theta)$\\
			Update model: $\theta = \theta - \eta \nabla \theta$		
		}	
	}
\end{algorithm}

\section{Conclusion \& Future Work}

Conversational agents such as Cortana, Alexa and Siri are continuously working on increasing their capabilities by adding new domains. The support of a new domain includes the design and development of a number of NLU components for domain classification, intents classification and slots tagging (including named entity recognition). Each component only performs well when trained on a large amount of labeled data. Second, these components are deployed on limited-memory devices which requires some model compression. Third, for some domains such as the health domain, it is hard to find a single training data set that covers all the required slot types. In this paper, we presented a multi-task transformer-based neural architecture for slot tagging that overcomes these mentioned problems and applied it into the biomedical domain (MT-BioNER). We formulated the training of a slot tagger using multiple data sets covering different slot types as a multi-task learning problem. We also reported the training and scoring time and compared it to the recent advancements. The experimental results on the biomedical domain have shown that MT-BioNER model trained using the proposed approach outperforms the previous state-of-the-art systems for slot tagging on different benchmark biomedical datasets in terms of both (time and memory efficiency) and effectiveness. The model has a shorter training time and inference time, and smaller memory footprint compared to the baseline multiple single-task based models. This is another advantage of our approach which can play an important role in the real-world applications. We run extra experiments to study the impact of fine-tuning the shared language model as it is found as a crucial point of our approach. Utilizing BERT base model and its comparison with BioBERT showed an interesting result in which for the domains that there is not a specific in-domain BERT model, the base model would also perform well with considering a penalty of few percent in the F1-score and more training iterations. Furthermore, we have showed through detailed analysis that the training algorithm plays a very important role in achieving the strong performance. The output slot tagger can be used by the conversational agent to better identify entities in the input utterances.\\

Lastly, we highlight several future directions to improve the multi-task BioNER model. Based on MT-BioNER results on three and four datasets, deep diving into the effect of dataset/entity types on the multi-task learning approaches could be an interesting future work. Overall this study, shows that how the named entity recognition capabilities of different biomedical systems can be obtained by employing recent trends from deep learning and multi-task learning based research. Incorporating such techniques can help the researchers to overcome the limitation of extensively labeled training data that is really hard to get in the biomedical domain. 

As a future work, we want to further explore the impact of overlap between the datasets on the overall model performance. Addition of the JNLPBA  to the training datasets results in degradation of the overall performance of the model. One possible reason for this degradation is that JNLPBA contains many genes and proteins which are represented as small abbreviations and code. These small abbreviations and genes can overlap with the entities in other datasets resulting in confusion and degradation of the overall model. We will like to explore ways to tackle overlap between entities that can degrade the model performance. We will also like to perform comparative analysis of different models on same input sentences to highlight the plus points of our model over other models.\\
Although, the datasets and experiments specified in this research paper are focused on the biomedical domain but the techniques and algorithms presented in this paper can be used in other domains and general conversational systems: ones that are not focused on the biomedical domain. We also want to analyze the performance of our NER model on general domain conversational systems in future work as well.
\bibliographystyle{ACM-Reference-Format}

\bibliography{nips-references}


\begin{thebibliography}{22}


\ifx \showCODEN    \undefined \def \showCODEN     #1{\unskip}     \fi
\ifx \showDOI      \undefined \def \showDOI       #1{#1}\fi
\ifx \showISBNx    \undefined \def \showISBNx     #1{\unskip}     \fi
\ifx \showISBNxiii \undefined \def \showISBNxiii  #1{\unskip}     \fi
\ifx \showISSN     \undefined \def \showISSN      #1{\unskip}     \fi
\ifx \showLCCN     \undefined \def \showLCCN      #1{\unskip}     \fi
\ifx \shownote     \undefined \def \shownote      #1{#1}          \fi
\ifx \showarticletitle \undefined \def \showarticletitle #1{#1}   \fi
\ifx \showURL      \undefined \def \showURL       {\relax}        \fi
\providecommand\bibfield[2]{#2}
\providecommand\bibinfo[2]{#2}
\providecommand\natexlab[1]{#1}
\providecommand\showeprint[2][]{arXiv:#2}

\bibitem[\protect\citeauthoryear{Ahmadvand, Sahijwani, Choi, and
  Agichtein}{Ahmadvand et~al\mbox{.}}{2019}]%
        {ahmadvand2019concet}
\bibfield{author}{\bibinfo{person}{Ali Ahmadvand}, \bibinfo{person}{Harshita
  Sahijwani}, \bibinfo{person}{Jason~Ingyu Choi}, {and} \bibinfo{person}{Eugene
  Agichtein}.} \bibinfo{year}{2019}\natexlab{}.
\newblock \showarticletitle{ConCET: Entity-aware topic classification for
  open-domain conversational agents}. In \bibinfo{booktitle}{\emph{Proceedings
  of the 28th ACM International Conference on Information and Knowledge
  Management}}. \bibinfo{pages}{1371--1380}.
\newblock


\bibitem[\protect\citeauthoryear{Alsentzer, Murphy, Boag, Weng, Jin, Naumann,
  and McDermott}{Alsentzer et~al\mbox{.}}{2019}]%
        {alsentzer2019publicly}
\bibfield{author}{\bibinfo{person}{Emily Alsentzer}, \bibinfo{person}{John~R
  Murphy}, \bibinfo{person}{Willie Boag}, \bibinfo{person}{Wei-Hung Weng},
  \bibinfo{person}{Di Jin}, \bibinfo{person}{Tristan Naumann}, {and}
  \bibinfo{person}{Matthew McDermott}.} \bibinfo{year}{2019}\natexlab{}.
\newblock \showarticletitle{Publicly available clinical BERT embeddings}.
\newblock \bibinfo{journal}{\emph{arXiv preprint arXiv:1904.03323}}
  (\bibinfo{year}{2019}).
\newblock


\bibitem[\protect\citeauthoryear{Cohn, Laish, Beryozkin, Li, Shafran, Szpektor,
  Hartman, Hassidim, and Matias}{Cohn et~al\mbox{.}}{2019}]%
        {cohn2017deid}
\bibfield{author}{\bibinfo{person}{Ido Cohn}, \bibinfo{person}{Itay Laish},
  \bibinfo{person}{Genady Beryozkin}, \bibinfo{person}{Gang Li},
  \bibinfo{person}{Izhak Shafran}, \bibinfo{person}{Idan Szpektor},
  \bibinfo{person}{Tzvika Hartman}, \bibinfo{person}{Avinatan Hassidim}, {and}
  \bibinfo{person}{Yossi Matias}.} \bibinfo{year}{2019}\natexlab{}.
\newblock \showarticletitle{Audio De-identification: A New Entity Recognition
  Task}.
\newblock
\urldef\tempurl%
\url{https://arxiv.org/pdf/1903.07037.pdf}
\showURL{%
\tempurl}


\bibitem[\protect\citeauthoryear{Collobert and Weston}{Collobert and
  Weston}{2008}]%
        {collobert2008unified}
\bibfield{author}{\bibinfo{person}{Ronan Collobert} {and}
  \bibinfo{person}{Jason Weston}.} \bibinfo{year}{2008}\natexlab{}.
\newblock \showarticletitle{A unified architecture for natural language
  processing: Deep neural networks with multitask learning}. In
  \bibinfo{booktitle}{\emph{Proceedings of the 25th international conference on
  Machine learning}}. ACM, \bibinfo{pages}{160--167}.
\newblock


\bibitem[\protect\citeauthoryear{Crichton, Pyysalo, Chiu, and
  Korhonen}{Crichton et~al\mbox{.}}{2017}]%
        {crichton2017neural}
\bibfield{author}{\bibinfo{person}{Gamal Crichton}, \bibinfo{person}{Sampo
  Pyysalo}, \bibinfo{person}{Billy Chiu}, {and} \bibinfo{person}{Anna
  Korhonen}.} \bibinfo{year}{2017}\natexlab{}.
\newblock \showarticletitle{A neural network multi-task learning approach to
  biomedical named entity recognition}.
\newblock \bibinfo{journal}{\emph{BMC bioinformatics}} \bibinfo{volume}{18},
  \bibinfo{number}{1} (\bibinfo{year}{2017}), \bibinfo{pages}{368}.
\newblock


\bibitem[\protect\citeauthoryear{Devlin, Chang, Lee, and Toutanova}{Devlin
  et~al\mbox{.}}{2018}]%
        {devlin2018bert}
\bibfield{author}{\bibinfo{person}{Jacob Devlin}, \bibinfo{person}{Ming-Wei
  Chang}, \bibinfo{person}{Kenton Lee}, {and} \bibinfo{person}{Kristina
  Toutanova}.} \bibinfo{year}{2018}\natexlab{}.
\newblock \showarticletitle{Bert: Pre-training of deep bidirectional
  transformers for language understanding}.
\newblock \bibinfo{journal}{\emph{arXiv preprint arXiv:1810.04805}}
  (\bibinfo{year}{2018}).
\newblock


\bibitem[\protect\citeauthoryear{GuoDong and Jian}{GuoDong and Jian}{2004}]%
        {guodong2004exploring}
\bibfield{author}{\bibinfo{person}{Zhou GuoDong} {and} \bibinfo{person}{Su
  Jian}.} \bibinfo{year}{2004}\natexlab{}.
\newblock \showarticletitle{Exploring deep knowledge resources in biomedical
  name recognition}. In \bibinfo{booktitle}{\emph{Proceedings of the
  International Joint Workshop on Natural Language Processing in Biomedicine
  and its Applications}}. Association for Computational Linguistics,
  \bibinfo{pages}{96--99}.
\newblock


\bibitem[\protect\citeauthoryear{Habibi, Weber, Neves, Wiegandt, and
  Leser}{Habibi et~al\mbox{.}}{2017}]%
        {habibi2017deep}
\bibfield{author}{\bibinfo{person}{Maryam Habibi}, \bibinfo{person}{Leon
  Weber}, \bibinfo{person}{Mariana Neves}, \bibinfo{person}{David~Luis
  Wiegandt}, {and} \bibinfo{person}{Ulf Leser}.}
  \bibinfo{year}{2017}\natexlab{}.
\newblock \showarticletitle{Deep learning with word embeddings improves
  biomedical named entity recognition}.
\newblock \bibinfo{journal}{\emph{Bioinformatics}} \bibinfo{volume}{33},
  \bibinfo{number}{14} (\bibinfo{year}{2017}), \bibinfo{pages}{i37--i48}.
\newblock


\bibitem[\protect\citeauthoryear{Huang, Altosaar, and Ranganath}{Huang
  et~al\mbox{.}}{2019}]%
        {huang2019clinicalbert}
\bibfield{author}{\bibinfo{person}{Kexin Huang}, \bibinfo{person}{Jaan
  Altosaar}, {and} \bibinfo{person}{Rajesh Ranganath}.}
  \bibinfo{year}{2019}\natexlab{}.
\newblock \showarticletitle{ClinicalBERT: Modeling Clinical Notes and
  Predicting Hospital Readmission}.
\newblock \bibinfo{journal}{\emph{arXiv preprint arXiv:1904.05342}}
  (\bibinfo{year}{2019}).
\newblock


\bibitem[\protect\citeauthoryear{Leaman and Lu}{Leaman and Lu}{2016}]%
        {leaman2016taggerone}
\bibfield{author}{\bibinfo{person}{Robert Leaman} {and}
  \bibinfo{person}{Zhiyong Lu}.} \bibinfo{year}{2016}\natexlab{}.
\newblock \showarticletitle{TaggerOne: joint named entity recognition and
  normalization with semi-Markov Models}.
\newblock \bibinfo{journal}{\emph{Bioinformatics}} \bibinfo{volume}{32},
  \bibinfo{number}{18} (\bibinfo{year}{2016}), \bibinfo{pages}{2839--2846}.
\newblock


\bibitem[\protect\citeauthoryear{Lee, Yoon, Kim, Kim, Kim, So, and Kang}{Lee
  et~al\mbox{.}}{2019}]%
        {lee2019biobert}
\bibfield{author}{\bibinfo{person}{Jinhyuk Lee}, \bibinfo{person}{Wonjin Yoon},
  \bibinfo{person}{Sungdong Kim}, \bibinfo{person}{Donghyeon Kim},
  \bibinfo{person}{Sunkyu Kim}, \bibinfo{person}{Chan~Ho So}, {and}
  \bibinfo{person}{Jaewoo Kang}.} \bibinfo{year}{2019}\natexlab{}.
\newblock \showarticletitle{Biobert: pre-trained biomedical language
  representation model for biomedical text mining}.
\newblock \bibinfo{journal}{\emph{arXiv preprint arXiv:1901.08746}}
  (\bibinfo{year}{2019}).
\newblock


\bibitem[\protect\citeauthoryear{Li, Chen, Chen, and Tang}{Li
  et~al\mbox{.}}{2015}]%
        {li2015hitsz_cdr}
\bibfield{author}{\bibinfo{person}{Haodi Li}, \bibinfo{person}{Qingcai Chen},
  \bibinfo{person}{Kai Chen}, {and} \bibinfo{person}{Buzhou Tang}.}
  \bibinfo{year}{2015}\natexlab{}.
\newblock \showarticletitle{HITSZ\_CDR System for Disease and Chemical Named
  Entity Recognition and Relation Extraction}. In
  \bibinfo{booktitle}{\emph{Proceedings of the Fifth BioCreative Challenge
  Evaluation Workshop. Sevilla: The fifth BioCreative challenge evaluation
  workshop}}, Vol.~\bibinfo{volume}{2015}. \bibinfo{pages}{196--201}.
\newblock


\bibitem[\protect\citeauthoryear{Liu, He, Chen, and Gao}{Liu
  et~al\mbox{.}}{2019}]%
        {liu2019multi}
\bibfield{author}{\bibinfo{person}{Xiaodong Liu}, \bibinfo{person}{Pengcheng
  He}, \bibinfo{person}{Weizhu Chen}, {and} \bibinfo{person}{Jianfeng Gao}.}
  \bibinfo{year}{2019}\natexlab{}.
\newblock \showarticletitle{Multi-task deep neural networks for natural
  language understanding}.
\newblock \bibinfo{journal}{\emph{arXiv preprint arXiv:1901.11504}}
  (\bibinfo{year}{2019}).
\newblock


\bibitem[\protect\citeauthoryear{Mikolov, Sutskever, Chen, Corrado, and
  Dean}{Mikolov et~al\mbox{.}}{2013}]%
        {mikolov2013distributed}
\bibfield{author}{\bibinfo{person}{Tomas Mikolov}, \bibinfo{person}{Ilya
  Sutskever}, \bibinfo{person}{Kai Chen}, \bibinfo{person}{Greg~S Corrado},
  {and} \bibinfo{person}{Jeff Dean}.} \bibinfo{year}{2013}\natexlab{}.
\newblock \showarticletitle{Distributed representations of words and phrases
  and their compositionality}. In \bibinfo{booktitle}{\emph{Advances in neural
  information processing systems}}. \bibinfo{pages}{3111--3119}.
\newblock


\bibitem[\protect\citeauthoryear{Peters, Neumann, Iyyer, Gardner, Clark, Lee,
  and Zettlemoyer}{Peters et~al\mbox{.}}{2018}]%
        {peters2018deep}
\bibfield{author}{\bibinfo{person}{Matthew~E Peters}, \bibinfo{person}{Mark
  Neumann}, \bibinfo{person}{Mohit Iyyer}, \bibinfo{person}{Matt Gardner},
  \bibinfo{person}{Christopher Clark}, \bibinfo{person}{Kenton Lee}, {and}
  \bibinfo{person}{Luke Zettlemoyer}.} \bibinfo{year}{2018}\natexlab{}.
\newblock \showarticletitle{Deep contextualized word representations}.
\newblock \bibinfo{journal}{\emph{arXiv preprint arXiv:1802.05365}}
  (\bibinfo{year}{2018}).
\newblock


\bibitem[\protect\citeauthoryear{Pyysalo, Ginter, Moen, Salakoski, and
  Ananiadou}{Pyysalo et~al\mbox{.}}{2013}]%
        {Pyysalo:2013b}
\bibfield{author}{\bibinfo{person}{S. Pyysalo}, \bibinfo{person}{F. Ginter},
  \bibinfo{person}{H. Moen}, \bibinfo{person}{T. Salakoski}, {and}
  \bibinfo{person}{S. Ananiadou}.} \bibinfo{year}{2013}\natexlab{}.
\newblock \showarticletitle{Distributional Semantics Resources for Biomedical
  Text Processing}. In \bibinfo{booktitle}{\emph{Proceedings of LBM 2013}}.
  \bibinfo{pages}{39--44}.
\newblock
\urldef\tempurl%
\url{http://lbm2013.biopathway.org/lbm2013proceedings.pdf}
\showURL{%
\tempurl}


\bibitem[\protect\citeauthoryear{Radford, Narasimhan, Salimans, and
  Sutskever}{Radford et~al\mbox{.}}{2018}]%
        {radford2018improving}
\bibfield{author}{\bibinfo{person}{Alec Radford}, \bibinfo{person}{Karthik
  Narasimhan}, \bibinfo{person}{Tim Salimans}, {and} \bibinfo{person}{Ilya
  Sutskever}.} \bibinfo{year}{2018}\natexlab{}.
\newblock \showarticletitle{Improving language understanding by generative
  pre-training}.
\newblock \bibinfo{journal}{\emph{URL https://s3-us-west-2. amazonaws.
  com/openai-assets/researchcovers/languageunsupervised/language understanding
  paper. pdf}} (\bibinfo{year}{2018}).
\newblock


\bibitem[\protect\citeauthoryear{Reshmi and Balakrishnan}{Reshmi and
  Balakrishnan}{2018}]%
        {reshmi2018enhancing}
\bibfield{author}{\bibinfo{person}{S Reshmi} {and} \bibinfo{person}{Kannan
  Balakrishnan}.} \bibinfo{year}{2018}\natexlab{}.
\newblock \showarticletitle{Enhancing Inquisitiveness of Chatbots Through NER
  Integration}. In \bibinfo{booktitle}{\emph{2018 International Conference on
  Data Science and Engineering (ICDSE)}}. IEEE, \bibinfo{pages}{1--5}.
\newblock


\bibitem[\protect\citeauthoryear{Sachan, Xie, Sachan, and Xing}{Sachan
  et~al\mbox{.}}{2017}]%
        {sachan2017effective}
\bibfield{author}{\bibinfo{person}{Devendra~Singh Sachan},
  \bibinfo{person}{Pengtao Xie}, \bibinfo{person}{Mrinmaya Sachan}, {and}
  \bibinfo{person}{Eric~P Xing}.} \bibinfo{year}{2017}\natexlab{}.
\newblock \showarticletitle{Effective use of bidirectional language modeling
  for transfer learning in biomedical named entity recognition}.
\newblock \bibinfo{journal}{\emph{arXiv preprint arXiv:1711.07908}}
  (\bibinfo{year}{2017}).
\newblock


\bibitem[\protect\citeauthoryear{UserTesting.com}{UserTesting.com}{2019}]%
        {chatbot_report}
\bibfield{author}{\bibinfo{person}{UserTesting.com}.}
  \bibinfo{year}{2019}\natexlab{}.
\newblock \bibinfo{title}{healthcare Chatbot diagnosis}.
\newblock
\newblock


\bibitem[\protect\citeauthoryear{Vaswani, Shazeer, Parmar, Uszkoreit, Jones,
  Gomez, Kaiser, and Polosukhin}{Vaswani et~al\mbox{.}}{2017}]%
        {vaswani2017attention}
\bibfield{author}{\bibinfo{person}{Ashish Vaswani}, \bibinfo{person}{Noam
  Shazeer}, \bibinfo{person}{Niki Parmar}, \bibinfo{person}{Jakob Uszkoreit},
  \bibinfo{person}{Llion Jones}, \bibinfo{person}{Aidan~N Gomez},
  \bibinfo{person}{{\L}ukasz Kaiser}, {and} \bibinfo{person}{Illia
  Polosukhin}.} \bibinfo{year}{2017}\natexlab{}.
\newblock \showarticletitle{Attention is all you need}. In
  \bibinfo{booktitle}{\emph{Advances in neural information processing
  systems}}. \bibinfo{pages}{5998--6008}.
\newblock


\bibitem[\protect\citeauthoryear{Wang, Zhang, Ren, Zhang, Zitnik, Shang,
  Langlotz, and Han}{Wang et~al\mbox{.}}{2018}]%
        {wang2018cross}
\bibfield{author}{\bibinfo{person}{Xuan Wang}, \bibinfo{person}{Yu Zhang},
  \bibinfo{person}{Xiang Ren}, \bibinfo{person}{Yuhao Zhang},
  \bibinfo{person}{Marinka Zitnik}, \bibinfo{person}{Jingbo Shang},
  \bibinfo{person}{Curtis Langlotz}, {and} \bibinfo{person}{Jiawei Han}.}
  \bibinfo{year}{2018}\natexlab{}.
\newblock \showarticletitle{Cross-type biomedical named entity recognition with
  deep multi-task learning}.
\newblock \bibinfo{journal}{\emph{Bioinformatics}} \bibinfo{volume}{35},
  \bibinfo{number}{10} (\bibinfo{year}{2018}), \bibinfo{pages}{1745--1752}.
\newblock


\end{thebibliography}

\end{document}